\documentclass[10pt,twocolumn,letterpaper]{article}

\usepackage[pagenumbers]{iccv} 
 
\definecolor{iccvblue}{rgb}{0.21,0.49,0.74}
\usepackage[pagebackref,breaklinks,colorlinks,allcolors=iccvblue]{hyperref}

\usepackage{multirow}

\def\paperID{7025} 
\def\confName{ICCV}
\def\confYear{2025}

\title{Open-set Cross Modal Generalization via Multimodal Unified Representation}

\author{Hai Huang$^{1,2}$ \quad Yan Xia$^{1}$ \quad Shulei Wang$^{1}$\quad Hanting Wang$^{1}$\\ \quad Minghui Fang$^{1}$ \quad Shengpeng Ji$^{1}$ \quad Sashuai Zhou$^{1}$  \quad Tao Jin$^{1}$ \quad Zhou Zhao$^{1,2\dagger}$
\\
$^{1}$ Zhejiang University
\quad$^{2}$ Shanghai Artificial Intelligence Laboratory\\
{\tt\small haihuangcode@outlook.com}\quad
{\tt\small zhaozhou@zju.edu.cn}
}

\usepackage[misc]{ifsym}
\newcommand\blfootnote[1]{
    \begingroup
    \renewcommand\thefootnote{}\footnote{#1}
    \addtocounter{footnote}{-1}
    \endgroup
}
\usepackage[hang]{footmisc}

\begin{document}
\maketitle
{
    \blfootnote{$^\dagger$Corresponding author}
}
\begin{abstract}

This paper extends Cross Modal Generalization (CMG) to open-set environments by proposing the more challenging Open-set Cross Modal Generalization (\textbf{OSCMG}) task. This task evaluates multimodal unified representations in open-set conditions, addressing the limitations of prior closed-set cross-modal evaluations. OSCMG requires not only cross-modal knowledge transfer but also robust generalization to unseen classes within new modalities, a scenario frequently encountered in real-world applications. Existing multimodal unified representation work lacks consideration for open-set environments. To tackle this, we propose \textbf{MICU}, comprising two key components: Fine-Coarse \textbf{M}asked multimodal \textbf{I}nfoNCE (FCMI) and \textbf{C}ross modal \textbf{U}nified Jigsaw Puzzles (CUJP). FCMI enhances multimodal alignment by applying contrastive learning at both holistic semantic and temporal levels, incorporating masking to enhance generalization. CUJP enhances feature diversity and model uncertainty by integrating modality-agnostic feature selection with self-supervised learning, thereby strengthening the model’s ability to handle unknown categories in open-set tasks. Extensive experiments on CMG and the newly proposed OSCMG validate the effectiveness of our approach. The code is available at \href{https://github.com/haihuangcode/CMG}{https://github.com/haihuangcode/CMG}.


\end{abstract}


\section{Introduction}
\label{sec:intro}
To address the challenge of scarce annotated data in downstream tasks involving rare modalities (e.g., point clouds, EEG signals), Cross Modal Generalization (CMG)~\cite{xia2024achieving} has been introduced as a novel task. This paradigm aims to establish unified representations through fine-grained pretraining on large-scale paired multimodal datasets, mapping semantically equivalent information across different modalities into a shared discrete dictionary. This framework enables zero-shot transfer of knowledge and capabilities learned from common modalities (such as images and text) to rare modalities in downstream applications, without requiring additional modality-specific annotations.

The method proposed by Xia~\etal~\cite{xia2024achieving} has achieved promising fine-grained semantic alignment results through feature disentangling and cross-modal contrastive prediction. However, their work relies on a closed-set assumption, where training and test classes remain consistent across tasks. In practical applications, the target modality for transfer often includes categories that do not exactly match those in the source domain. Directly applying previous method~\cite{duan2022multi,lu2022unified,liu2021cross,sarkar2022xkd,zhao2022towards,xia2024achieving} for cross-modal generalization would lead to misclassification of these unknown categories, limiting its applicability in real-world scenarios.

\begin{figure}
    \centering
    \includegraphics[width=0.80\linewidth]{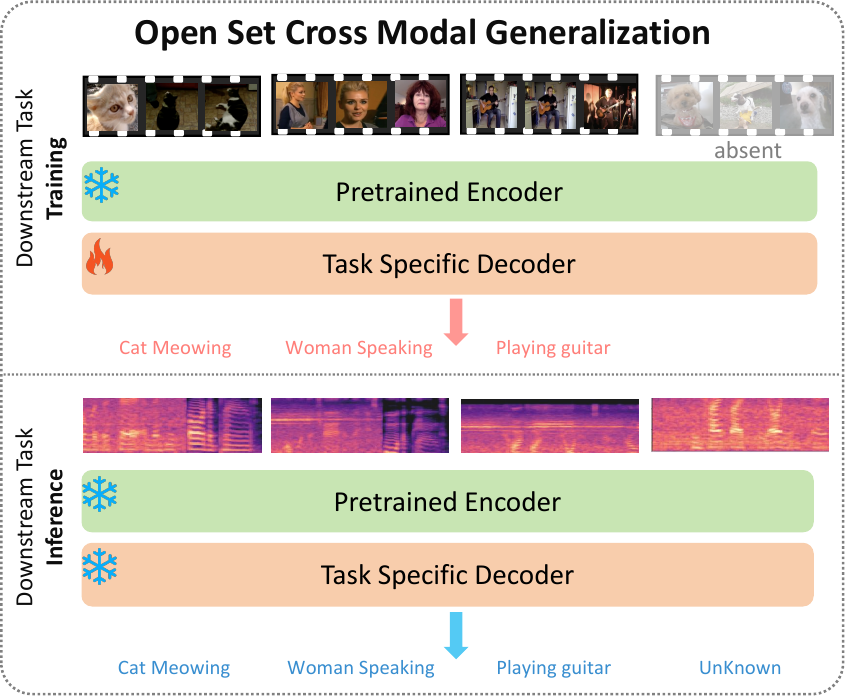}
    \vspace{-2mm}
    \caption{After unsupervised pretraining, the model is directly transferred to unseen modalities and unseen categories in downstream tasks.}
    \label{fig:oscmg_illu}
    \vspace{-5mm}
\end{figure}


Therefore, we introduce the \textbf{Open-Set Cross-Modal Generalization (OSCMG)} task, designed to enhance models' cross-modal generalization capabilities in open-set environments. The OSCMG task requires models not only to achieve unified representations across different modalities but also to ensure that these representations are highly generalizable, enabling effective distinction between known and unknown classes. Specifically, this approach pretrains the model in an unsupervised setting, then fine-tunes it on downstream tasks with modality $a$, containing only the class set $V$, and enables it to generalize to modality $b$, which includes a broader class set $U$, where $V \subset U$; a graphical depiction can be seen in Figure~\ref{fig:oscmg_illu}. Similar open-set tasks include Open-Set Domain Generalization (OSDG)\cite{shu2021open} and Multimodal Open-Set Domain Generalization (MM-OSDG)\cite{dong2024towards}, which extend the challenges of Domain Generalization (DG)\cite{carlucci2019domain} and Multimodal Domain Generalization (MMDG)\cite{dong2024simmmdg} to open-domain scenarios, with specific differences outlined in~\cref{tab:tasks setting}.

\begin{table}[t]
\centering
\begin{tabular}{cccc}
\toprule
$C_{s}\neq C_{t}$ & Multimodal & $M_{s}\neq M_{t}$ & {\bf Task Setting}\\
\midrule
- & - & - & DG~\cite{carlucci2019domain}\\
\checkmark & - & - & OSDG~\cite{shu2021open}\\
- & \checkmark & - & MMDG~\cite{dong2024simmmdg}\\
\checkmark & \checkmark & - & MM-OSDG~\cite{dong2024towards}\\
- & \checkmark & \checkmark & CMG~\cite{xia2024achieving}\\
\checkmark & \checkmark & \checkmark & OSCMG\\
\bottomrule
\end{tabular}%
\vspace{-2mm}
\caption{The differences between OSCMG and other related tasks. $M_{s}$ and $M_{t}$ represent the source and target modalities, while $C_{s}$ and $C_{t}$ represent the labels of the source and target modalities.}
\label{tab:tasks setting}
\vspace{-5mm}
\end{table}


As a novel task, OSCMG primarily encompasses two challenging aspects. \textbf{(1)} To achieve cross-modal generalization, it is crucial to establish effective multimodal unified representations. However, previous works have predominantly focused on alignment at a singular level. For instance, methods like CLIP~\cite{radford2021learning} and ImageBind~\cite{girdhar2023imagebind} perform coarse-grained alignment by average pooling features from different modalities, which can easily overlook fine-grained cross-modal alignment relationships. Xia~\cite{xia2024achieving} addresses the challenge of fine-grained multimodal alignment through cross-modal contrastive predictive coding. Nonetheless, it tends to overlook the holistic semantic associations between different modalities. \textbf{(2)} Since large-scale labeled multimodal data is difficult to obtain, the construction of a unified representation primarily relies on learning from vast amounts of unlabeled multimodal data. We propose OSCMG to evaluate the performance of unified representation under more challenging conditions, thus adopting an unsupervised setting. This setting renders most existing label-dependent OSDG methods, such as DAML~\cite{shu2021open} and MEDIC~\cite{wang2023generalizable}, inapplicable to OSCMG. In contrast, MMJP~\cite{dong2024towards}, designed as a self-supervised learning approach that does not require label information, was proposed to tackle the MM-OSDG challenge and has demonstrated strong generalization capabilities in open-domain multimodal scenarios. However, MMJP is not suitable for the OSCMG task. Its core mechanism relies on utilizing all modalities to perform the jigsaw puzzles, leveraging cross-modal complementarity to enhance performance in MM-OSDG. This design makes MMJP highly sensitive to modality-specific semantics, as it depends on information from all modalities during training. However, such sensitivity can be detrimental to the learning of a unified representation, as it emphasizes modality-specific features that may negatively impact the representation’s generalization~\cite{xia2024achieving}.


To address these challenges, we propose MICU, a novel approach that combines strong generalization with enhanced multimodal alignment through two key components: Fine-Coarse \textbf{M}asked Multimodal \textbf{I}nfoNCE (FCMI) and \textbf{C}ross-modal \textbf{U}nified Jigsaw Puzzles (CUJP). \textbf{(1)} FCMI refines and strengthens multimodal alignment by applying masked contrastive learning at both inter-sample (holistic semantic) and intra-sample (temporal) levels, thereby capturing broad semantic consistency and fine-grained alignment to construct a more effective multimodal unified representation space. \textbf{(2)} Considering the unsupervised setting of OSCMG pre-training, we adopt a self-supervised learning approach that does not require label information. However, as previously mentioned, MMJP~\cite{dong2024towards} is highly sensitive to modality-specific semantic features, whereas OSCMG aims to learn a unified representation by minimizing the influence of modality-specific information. To address this, we propose CUJP, which disregards modality distinctions and treats all modalities as a single unified modality. During the jigsaw puzzle process, CUJP randomly selects feature split blocks from any modality, enabling modal-agnostic learning. Furthermore, benefiting from the partitioning mechanism of the jigsaw puzzle, CUJP achieves finer-grained alignment compared to previous unified representation approaches that primarily focus on aligning entire samples~\cite{liu2021cross,zhao2022towards,xia2024achieving}, ensuring consistency at the block level. Additionally, CUJP significantly reduces computational complexity compared to MMJP, as it does not require using all feature blocks from every modality. For instance, in a three-modality setting where each modality's features are split into four parts, MMJP requires $12!=479001600$ sorting computations, whereas CUJP only requires $4!=24$, leading to a substantial improvement in computational efficiency.  Our contributions can be summarized as follows:
\begin{itemize}
\item We propose {\bf OSCMG}, which enables the evaluation of multimodal unified representations under more realistic and complex challenges. This approach evaluates the model's ability not only to generalize across modalities but also to transfer knowledge to unseen categories.
\item We propose {\bf MICU}, which comprises {\bf FCMI} and {\bf CUJP}. FCMI achieves multimodal alignment through fine- and coarse-grained contrastive learning across temporal and holistic semantic levels, enhanced by a masking mechanism. CUJP enhances modality-agnostic performance by integrating discrete unified representations with a jigsaw puzzle approach, splitting and randomly rearranging the quantized representations.
\item Our model achieves state-of-the-art performance on both CMG and OSCMG tasks, demonstrating the effectiveness of the proposed methods.
\end{itemize}

\section{Related Work}
\label{sec:related work}

{\bf Multimodal Unified Representation.}
Recent efforts in multimodal unified representation focus on aligning different modalities in a shared latent space~\cite{petridis2018audio,sarkar2022xkd,andonian2022robust}, training modal-general encoders for cross-modal feature extraction~\cite{chen2020uniter,wang2022vlmixer}, and using cross-modal knowledge distillation to facilitate information transfer between modalities~\cite{sarkar2022xkd,pedersoli2022estimating}. Bridging techniques have also been proposed to connect continuous representation spaces to leverage complementary strengths~\cite{zhang2024extending}. To improve interpretability, codebooks or prototypes are used for unified representations, mapping multimodal features into discrete forms~\cite{duan2022multi,lu2022unified,liu2021cross,zhao2022towards,xia2024achieving,huang2024enhancing,huang2025bridging,huang2025semantic}. For instance, Duan \etal~\cite{duan2022multi} uses Optimal Transport to align features with prototypes, while Zhao \etal~\cite{zhao2022towards} enhances mutual information via self-cross-reconstruction. Xia \etal~\cite{xia2024achieving} addresses imperfect alignment by mapping sequences into a common discrete space. We retained the consideration that paired multimodal data may not be perfectly aligned and proposed FCMI, which is easier to train compared to decoupling-based methods. Furthermore, we combined the highly effective Jigsaw Puzzle approach from the self-supervised learning domain with discrete representations, introducing CUJP, which achieves better unified representation performance.

\begin{figure*}[t]
  \centering
   \includegraphics[width=0.95\linewidth]{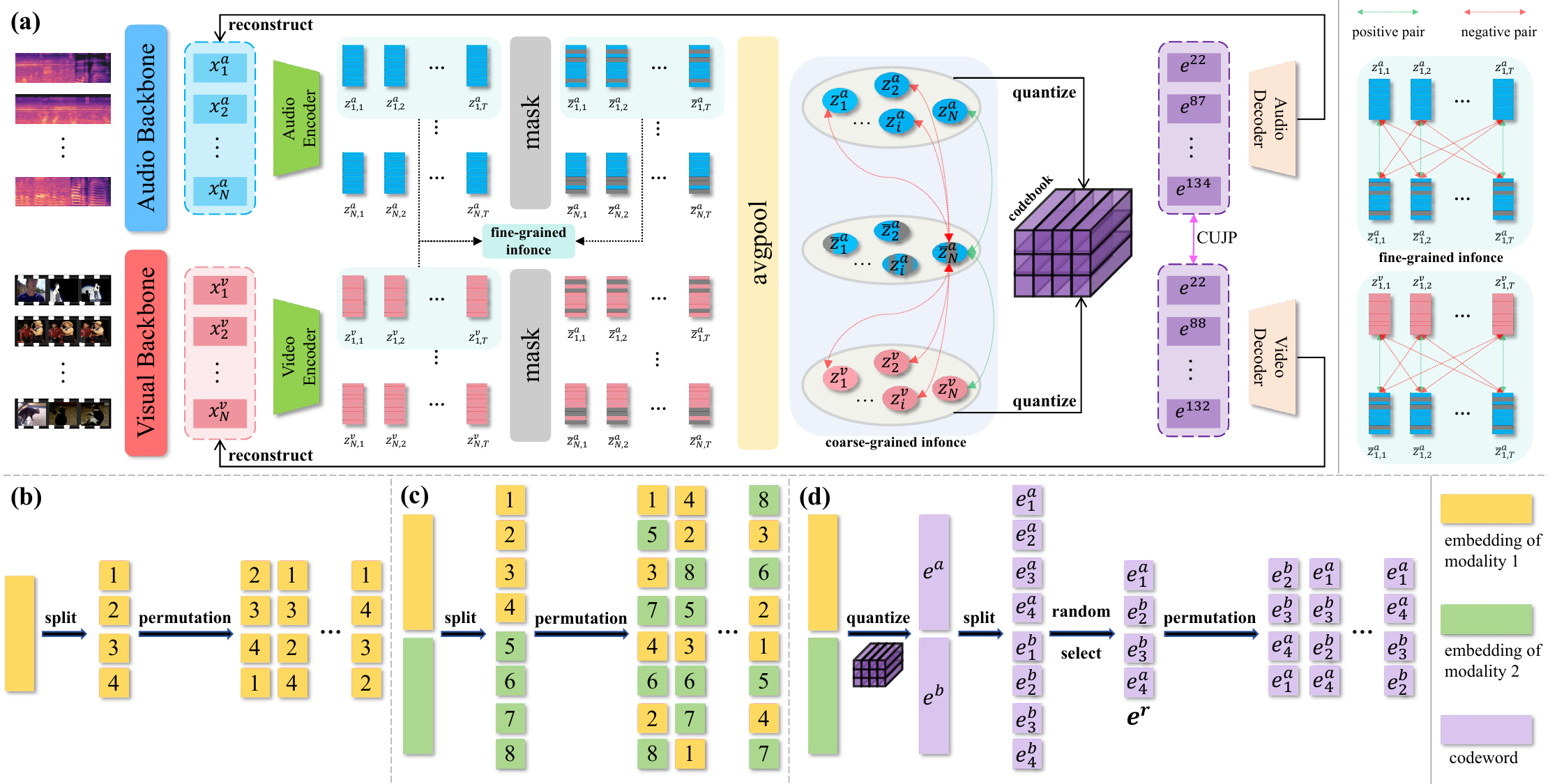}
   \caption{(a) The architecture of MICU, illustrated with an example of fine and coarse InfoNCE with masked audio and video, as well as with masked audio and audio. (b) Single-modal Jigsaw Puzzles. (c) Multimodal Jigsaw Puzzles. (d) Our proposed Cross modal Unified Jigsaw Puzzles.
   \label{fig:framework}
   }
\vspace{-3mm}
\end{figure*}

\noindent
{\bf Domain and Cross-Modal Generalization.} DG has been instrumental in enabling models to generalize to unseen target domains without direct access to target domain data, and has found applications in diverse fields such as medical imaging~\cite{li2020domain,liu2021feddg} and action recognition~\cite{planamente2022domain}. Common DG methods include feature representation learning~\cite{tzeng2014deep,ganin2016domain,pan2018two}, data augmentation~\cite{tobin2017domain,zhang2018mixup}, and domain-agnostic learning strategies such as domain adversarial learning~\cite{ganin2016domain, 2020-6-1119, 2020-4-608} and meta-learning~\cite{li2018learning} to handle domain shifts. As multimodal research has advanced, MMDG ~\cite{planamente2022domain,dong2024simmmdg} emerged to address the additional complexity of generalizing across different modalities. 


In scenarios where the target domain may include categories unseen during training, OSDG\cite{shu2021open} addresses both domain generalization and unknown class detection. This concept has further developed into MM-OSDG\cite{dong2024towards}, with tasks such as MOOSA leveraging multimodal self-supervised learning to enhance generalization and open-set recognition in multimodal contexts. Similarly, CMG, like MMDG, faces challenges in open-set environments. To bridge this gap in evaluating multimodal unified representations, we propose the Open-set Cross-Modal Generalization  (OSCMG) task, which requires models to transfer knowledge across modalities and adapt to unseen classes within new modalities.

\section{Method}
In this section, we first provide a detailed definition of the proposed OSCMG task, followed by an introduction to our new architecture, MICU, designed to address this challenge. MICU primarily integrates the concepts of masked contrastive learning and self-supervised learning. We will introduce its two constituent modules separately, whereas Figure~\ref{fig:framework} illustrates the overall model architecture.

\subsection{Open-set Cross Modal Generalization}
OSCMG shares the same pre-training setup as CMG, where multimodal data is learned in an unsupervised manner to obtain a unified multimodal representation. The key difference lies in the evaluation of downstream tasks, OSCMG is designed to assess a model’s cross-modal generalization ability under open-set conditions. Specifically, it evaluates the model's capacity to transfer knowledge from a source modality to a target modality while handling unseen classes absent in the source modality. During training, the model is trained on a source modality $M_s$ and tested on a target modality $M_t$, where the class set of the source modality $V$ is a subset of the class set $U$ in the target modality, i.e., $V \subset U$. This setup challenges the model to generalize across modalities while also adapting to novel categories not encountered during training, providing a more comprehensive evaluation of cross-modal learning capabilities.


During training, the model learns representations for inputs from a source modality using the encoder $\Phi^{M_s}$ and the downstream decoder $\mathbf{D}$. The process is formulated as follows:
\begin{equation}
\mathbf{E}(\mathbf{D}(\Phi^{M_s}(\mathbf{x}^{M_s}_{i})), \mathbf{y}^{M_s}_{i}).
\end{equation}
where $\mathbf{x}^{M_s}_{i}$ is the input, $\mathbf{y}^{M_s}_{i}$ is the corresponding label, and $\mathbf{E}$ denotes the evaluation function. In the testing phase, the model is evaluated on a different target modality $M_t$, assessing its generalization capability:
\begin{equation}
\mathbf{E}(\mathbf{D}(\Phi^{M_t}(\mathbf{x}^{M_t}_{i})), \mathbf{y}^{M_t}_{i}).
\end{equation}
The parameters of the encoders $\Phi^{M_s}$ and $\Phi^{M_t}$ remain frozen during both downstream training and testing, as they are fully determined during the pre-training process. With only the parameters of the decoder $\mathbf{D}$ being updated during downstream training. Additionally, the encoders are derived from a multimodal model pretrained in an unsupervised manner, while the decoder varies according to the downstream task, typically implemented as a linear probe.

\subsection{Fine-Coarse Masked Multimodal InfoNCE}
In the field of multimodal unified representation, contrastive learning is a widely used alignment method. Liu \etal~\cite{liu2021cross} enhanced discrete representations through contrastive learning, significantly improving unified representation performance, while Xia \etal~\cite{xia2024achieving} incorporated cross-modal contrastive learning into their disentanglement framework. Building on these foundations, we introduce \textbf{FCMI}, an improved InfoNCE approach designed for multimodal unified representation. FCMI strengthens alignment by applying contrastive learning at both inter-sample (holistic semantic) and intra-sample (temporal) levels, ensuring both broad semantic consistency and fine-grained alignment. To enhance model generalization, we introduce masking within contrastive learning, inspired by SemSeg~\cite{yang2024class}, which builds class embeddings to recognize unknown categories, and Mask2Anomaly~\cite{rai2024mask2anomaly}, which uses masked contrastive learning to sharpen the boundary between known and anomalous classes. 

As shown in Figure~\ref{fig:framework}(a), FCMI is divided into two parts: fine-grained and coarse-grained masked contrastive learning. The paired features extracted by the backbone from each modality are denoted as $\{(\mathbf{x}^a_i,\mathbf{x}^b_i)\}$, where $\{a,b\}$ representing paired modals. For each modality, an encoder $\Phi^m$, where $m \in \{a,b\}$, is introduced to map the features to a uniform feature size $\mathbf{z}^{m}_{i} \in \mathbb{R}^{T \times D}$, where $T$ and $D$ represent the audio-video time dimension and the latent feature dimension, respectively:

\begin{equation}
    \mathbf{z}^{m}_{i} = \Phi^m(\mathbf{x}^{m}_{i}), \ m \in \{a,b\}.
\label{equ1} 
\end{equation}

We then apply a mask to the features, resulting in $\mathbf{\Bar{z}}^{m}_{i} = Mask(\mathbf{z}^{m}_{i})$. This masking is sample-specific, meaning the mask is consistent across different timesteps for the same sample. To ensure effective cross-modal masked contrastive learning, the masked positions are aligned across corresponding samples' different modalities, which will be discussed further in Figure~\ref{fig:ablation_mask}.

The fine-grained masked contrastive learning is applied to different timesteps of a single sample pair. The masked features at a specific timestep are contrasted with the unmasked features of the corresponding timestep from other modalities as positive pairs, while the remaining timesteps serve as negative pairs.

\begin{equation}
\begin{aligned}
L_\text{fine} = -\frac{1}{N} \frac{1}{T}  \sum_{i=1}^N \sum_{j=1}^T 
&\log \left[  \frac{\exp(\mathbf{\Bar{z}}^{m}_{i,j}\cdot (\mathbf{z}^{n}_{i,j})^{\top}/\tau)}{\sum_{k=1}^T \exp(\mathbf{\Bar{z}}^{m}_{i,j}\cdot (\mathbf{z}^{n}_{i,k})^{\top}/\tau)} \right], \\
& m,n \in \{a,b\}, 
\end{aligned}
\label{equ2} 
\end{equation} where $N$ represents the number of samples, $\top$ denotes transpose, and $\tau$ is the temperature parameter. Both $m$ and $n$ can represent the same modality, allowing for cross-modal as well as intra-modal alignment. This loss enables the model to learn fine-grained cross-modal alignment. Adjacent modalities time steps can serve as hard negatives, a strategy that effectively enhances contrastive learning by enforcing finer temporal discrimination and improving robustness.

Simultaneously, coarse-grained masked contrastive learning is applied across samples, where the masked features of a single sample are contrasted with the corresponding complete features from other modalities as positive pairs, and other samples as negative pairs.

\begin{equation}
\begin{aligned}
L_\text{coarse} = -\frac{1}{N}\sum_{i=1}^N
&\log \left[  \frac{\exp(\mathbf{\Bar{z}}^{m}_{i}\cdot (\mathbf{z}^{n}_{i})^{\top}/\tau)}{\sum_{j=1}^N \exp(\mathbf{\Bar{z}}^{m}_{i}\cdot (\mathbf{z}^{n}_{j})^{\top}/\tau)} \right], \\
& m,n \in \{a,b\}, 
\end{aligned}
\label{equ3} 
\end{equation} this loss facilitates the learning of multimodal alignment at the holistic semantic level. 

The cross-modal InfoNCE between unmasked features is not applied, as indirect alignment has already been achieved through modality masking. Adding this extra computation would not significantly improve the results.

\subsection{Cross Modal Unified Jigsaw Puzzle}
Previous studies~\cite{noroozi2016unsupervised,carlucci2019domain} have used Jigsaw puzzles to learn visual representations, where the task is to reconstruct an original image from shuffled parts. MMJP~\cite{dong2024towards} extended this idea to MM-OSDG. While CUJP shares the use of Jigsaw puzzles with MMJP, it differs by operating on unified discrete representations rather than shuffling all modality parts. Specifically, CUJP utilizes quantized features $\hat{\mathbf{z}}^{m}_{i,t}$ from the codebook, where each segment is a randomly selected codeword $e$ from any modality. This design significantly enhances modality-agnostic feature diversity and uncertainty, making CUJP particularly well-suited for OSCMG. It effectively integrates the advantages of MMJP in open-domain multimodal learning while preserving the unified representation property, which does not require modality-specific information. The illustrations of the three different Jigsaw puzzles are shown in subfigures (b), (c), and (d) of Figure~\ref{fig:framework}.


To explicitly represent the unified representation of different modalities, we utilize a shared latent codebook $\mathbf{E} \in \mathbb{R}^{H\times D}$ across multi modalities. We apply a vector quantization $VQ$ operation to map the multimodal features $\mathbf{z}^{a}_{i}$ and $\mathbf{z}^{b}_{i}$ into discrete latent codes. Here, $t \in [0, T)$, and $T$, $H$, and $D$ represent the time steps, the size of the discrete latent space, and the hidden dimension, respectively. 
\begin{equation}
\begin{split}
    \hat{\mathbf{z}}^{m}_{i,t} &= VQ(\Phi^{m}(\mathbf{x}^{m}_{i,t})) = VQ(\mathbf{z}^{m}_{i,t}) = e_{l}, \\
     {\rm where} \ l &= argmin_{j}\lvert\lvert \Phi^{m}(x) - e_{j}\rvert\rvert_{2}, \ m \in \{a,b\}.
\end{split}
\end{equation}

Not all $\hat{\mathbf{z}}^{m}_{i,t}$ are utilized in the process, and each segment is treated as modality-agnostic, enhancing uncertainty to aid open-set detection. This contrasts with MMJP, which explicitly differentiates between modalities.

The modality codes are divided into $O$ segments of equal length: $e^a = [e^a_1, e^a_2, \dots, e^a_O]$ and $e^b = [e^b_1, e^b_2, \dots, e^b_O]$. These segments are randomly selected across modalities to form $e^r = [e^m_1, e^m_2, \dots, e^m_O]$, where $m \in \{a, b\}$. One possible permutation is $\tilde{e}^o = [e^{m_2}_2, e^{m_n}_O, \dots, e^{m_1}_1]$. The $O$ segments are subsequently shuffled to produce different permutations, yielding a total of $O!$ possible combinations. Among these, we randomly sample $P$ permutations and assign each a unique index to serve as its label.

An auxiliary classification task is introduced for each sample instance, formulated as $\{(\tilde{e} \in \tilde{e}^o, o)\}_{o=1}^{P}$, where $\tilde{e} \in \tilde{e}^o$ denotes the recomposed embeddings, and $o \in \{1, \ldots, P\}$ indicates the associated permutation index. The goal is to optimize the cross-modal jigsaw loss $L_{\text{cujp}}(\mathcal{H}(\tilde{e}), o)$, with $\mathcal{H}$ being the classifier used for recognizing the permutation, and $L_{\text{cujp}}$ denoting the conventional cross-entropy loss. Furthermore, as the combined feature dimension in CUJP matches that of a single modality, the number of required permutations is reduced, enhancing computational efficiency.

\subsection{Final Loss}
In addition to the previously mentioned losses, the following losses are also required:
\begin{equation}
\begin{split}
    &\underbrace{\|\mathbf{x}_{i}^{m} - D(\hat{\mathbf{e}}_{i}^{m})\|_2^2}_{L_{\text{recon}}} +\underbrace{ \|\Phi^{m}(\mathbf{x}_{i}^{m}) - \text{sg}[\mathbf{e}]\|_2^2}_{L_{\text{commit}}}
\end{split}
\end{equation}

Here, $\text{sg}$ denotes the stop-gradient operation. The reconstruction loss, $L_{\text{recon}}$, measures the difference between the outputs of each modality projector $\Phi^m$ and the original inputs using Mean Squared Error (MSE). The commitment loss, $L_{\text{commit}}$, computes the MSE between the encoder results and their quantized codes. In this work, we replace the traditional VQ loss with Exponential Moving Average (EMA), as EMA offers greater robustness. The final loss is as follows, $\lambda_{1}, \lambda_{2}, \lambda_{3}, \lambda_{4}$ are hyperparameters:
\begin{equation}
L=\lambda_{1} (L_{fine}+L_{coarse})+\lambda_{2}L_{cujp}+\lambda_{3}L_{recon}+\lambda_{4}L_{commit}
\end{equation}


\section{Experiment}

\begin{table*}[htbp]
    \centering
    \resizebox{1.0\textwidth}{!}{
    \begin{tabular}{ccccccccccccccccc}
    \toprule
    \multirow{3}{*}{Dataset} & \multirow{3}{*}{Method} 
                            & \multicolumn{7}{c}{Split1} & & \multicolumn{7}{c}{Split2} \\ \cline{3-9} \cline{11-17}
                            & & \multicolumn{3}{c}{V$\rightarrow$A} & & \multicolumn{3}{c}{A$\rightarrow$V} & & \multicolumn{3}{c}{V$\rightarrow$A} &  &\multicolumn{3}{c}{A$\rightarrow$V} \\ \cline{3-5} \cline{7-9} \cline{11-13} \cline{15-17}
                            & & OS* & UNK & \textbf{HOS} & & OS* &  UNK & \textbf{HOS} & & OS* & UNK & \textbf{HOS} & &OS* & UNK & \textbf{HOS} \\
                            \midrule
    \multirow{5}{*}{\bf AVE} &
    CODIS~\citep{duan2022multi} & 36.41 & 47.33 & 41.16 & & 26.31 & 37.29 & 30.85 & & 34.51 & 55.76 & 42.63 & & 27.71 & 52.41 & 36.25 \\ 
    &TURN~\citep{zhao2022towards} & 35.37 & 49.26 & 41.18 & & 27.13 & 39.41 & 32.14 & & 31.73 & 58.13 & 41.05 & & 25.89 & 56.26 & 35.46 \\ 
    &CMCM~\citep{liu2021cross} & 39.09 & 53.48 & 45.17 & & 30.21 & 45.93 & 36.45 & & 34.51 & 62.86 & 44.56 & & 30.78 & 61.31 & 40.98 \\ 
    &DCID~\citep{xia2024achieving} & 45.29 & 59.78 & 51.54 & & 34.98 & 42.46 & 38.36 & & 41.14 & 68.60 & 51.44 & & 34.18 & 67.44 & 45.37 \\ 
    &MICU & 51.57 & 57.54 & \textbf{54.39} & & 34.98 & 64.80 & \textbf{45.43} & & 47.15 & 79.07 & \textbf{59.08} & & 35.44 & 80.23 & \textbf{49.17} \\ \midrule
    \multirow{5}{*}{\bf UCF} &
    CODIS~\citep{duan2022multi} & 17.51 & 43.17 & 24.91 & & 23.66 & 49.04 & 31.92 & & 16.33 & 45.32 & 24.01 & & 17.80 & 43.78 & 25.31 \\ 
    &TURN~\citep{zhao2022towards} & 15.43 & 43.39 & 22.76 & & 22.05 & 53.75 & 31.27 & & 17.41 & 44.76 & 25.07 & & 18.43 & 44.96 & 26.14 \\ 
    &CMCM~\citep{liu2021cross} & 21.41 & 50.09 & 30.00 & & 25.38 & 51.63 & 34.03 & & 18.78 & 46.72 & 26.79 & & 21.67 & 47.87 & 29.83\\ 
    &DCID~\citep{xia2024achieving} & 25.08 & 55.06 & 34.46   &&  29.62 & 53.35 & 38.09 & & 18.52 & 58.97 & 28.18  &&   25.83 & 48.28 & 33.65 \\ 
    &MICU & 29.40 & 61.69 & \textbf{39.82} & &    27.48 & 72.96 & \textbf{39.92} & & 24.33 & 60.05 & \textbf{34.64}   & &  23.90 & 68.25 & \textbf{35.41} \\ \midrule
    \multirow{5}{*}{\bf UCF(v)$\leftrightarrow$VGG(a)} &
    CODIS~\citep{duan2022multi} & 62.75 & 75.35 & 68.48 & & 43.61 & 63.71 & 51.78 & & 47.71 & 79.16 & 59.54 & & 41.61 & 72.14 & 52.78 \\ 
    &TURN~\citep{zhao2022towards} & 59.73 & 78.52 & 67.85 & & 41.52 & 64.40 & 50.49 & & 51.31 & 75.53 & 61.11 & & 40.73 & 75.62 & 52.94 \\ 
    &CMCM~\citep{liu2021cross} & 68.44 & 77.17 & 72.54 & & 43.67 & 68.89 & 53.45 & & 50.17 & 84.62 & 62.99 & & 44.61 & 78.43 & 56.87 \\ 
    &DCID~\citep{xia2024achieving} & 79.16 & 88.53 & 83.58 & & 56.47 & 77.34 & 65.28 & & 54.97 & 95.83 & 69.87 & & 50.00 & 83.22 & \textbf{62.49} \\ 
    &MICU & 81.72 & 93.23 & \textbf{87.09} & & 68.71 & 70.70 & \textbf{69.69} & & 66.77 & 87.18 & \textbf{75.62} & & 47.43 & 86.13 & 61.17 \\ \bottomrule
    \end{tabular}
    }
    \caption{Comparison of our model with previous SOTA models on OSCMG. Split1 and Split2 represent different class partitioning schemes of the training set for each dataset, where Split1 corresponds to the scheme with fewer classes in the training set.}
    \label{tab:oscmg}
    \vspace{-3mm}
\end{table*}

\subsection{Experimental Setting}

\noindent
\textbf{Pretrain: }We use VGGsound-AVEL40K~\cite{chen2020vggsound,zhou2022contrastive} with text provided by~\cite{xia2024achieving} to train unified representation. 

\noindent
\textbf{Downstream: }We propose the OSCMG problem, which includes three tasks: classification on the AVE~\cite{avel} and UCF~\cite{soomro2012ucf101} datasets, and a cross-dataset classification task between UCF and VGG~\cite{chen2020vggsound} (UCF$\leftrightarrow$VGG). The AVE dataset originally contains 28 classes. We split the data based on the original labels into a 1:1 and 3:1 ratio, resulting in 14-class or 21-class training sets, which are then tested on the full 28 classes. For UCF, after filtering out classes without audio data from the original 101 classes, we obtained 51 classes. The data was split into training sets with either 17 or 34 classes in a 1:2 and 2:1 ratio, while testing was performed on the complete 51 classes. For UCF$\leftrightarrow$VGG, we filtered the labels to retain 16 common classes between UCF and VGG, splitting them into 1:1 and 3:1 ratios. This resulted in training sets with 8 or 12 classes, and testing was conducted on all 16 classes. It is important to note that some UCF classes do not have audio data, so in UCF$\leftrightarrow$VGG, we only use UCF's video modality (v) paired with VGGSound's audio modality (a).


The CMG problem includes four tasks: cross-modal classification on AVE~\cite{avel} and UCF$\leftrightarrow$VGG~\cite{soomro2012ucf101,chen2020vggsound}, and cross-modal localization tasks on AVVP~\cite{tian2020unified} and AVE$\rightarrow$AVVP. Additionally, we conducted experiments on cross-modal zero-shot retrieval.

\noindent
\textbf{Evaluation Metrics: }The evaluation metrics used in OSCMG are OS, UNK, and HOS, which have been widely adopted in prior open-set recognition works~\cite{bucci2020effectiveness,li2023adjustment,dong2024towards}. The HOS metric is calculated as $\text{HOS} = \frac{2 \times \text{OS}^* \times \text{UNK}}{\text{OS}^* + \text{UNK}}$, where $\text{OS}^*$ refers to the accuracy for known categories, and UNK corresponds to the accuracy for unknown categories. Unlike OS, HOS offers a more comprehensive performance measure by balancing results across known and unknown classes, which is crucial when accuracy for unknown classes is notably lower, underscoring the need for effective detection of unknown categories. For CMG, we employ different evaluation metrics depending on the task. Precision is used for classification tasks on AVE~\cite{avel}, VGG~\cite{zhou2022contrastive, zhou2021positive}, and UCF~\cite{soomro2012ucf101}, while the F1-score is utilized for localization tasks on AVVP~\cite{tian2020unified} and AVE$\rightarrow$AVVP. For cross-modal zero-shot retrieval~\cite{chen2011collecting, drossos2020clotho}, recall is the primary evaluation metric.

\noindent
\textbf{Implementation Details: }We compare our model against several state-of-the-art methods in multimodal unified discrete representations and multimodal domain generalization, including CODIS~\cite{duan2022multi}, TURN~\cite{zhao2022towards}, CMCM~\cite{liu2021cross}, and DCID~\cite{xia2024achieving}. These models are evaluated across our tasks and various downstream scenarios. For both $L_{\text{fine}}$ and $L_{\text{coarse}}$, the temperature parameter $\tau$ is set to 1.0, the mask ratio of FCMI is set to 30\%. All experiments, as shown in Tables~\ref{tab:oscmg},~\ref{tab:cmg},~\ref{tab:retrieval},~\ref{tab:ablation_module_oscmg},~\ref{tab:ablation_module_cmg}, and Figures~\ref{fig:codebook_visualization},~\ref{fig:ablation_jp},~\ref{fig:ablation_mask}, use a codebook size of 400 with an embedding dimension of 256. To ensure a fair comparison, all experiments, except those in Tables~\ref{tab:Enhanced_backbone},~\ref{tab:VAF}, follow the same backbone settings as DCID~\cite{xia2024achieving}. However, since DCID employs relatively outdated backbones for video and audio, we introduce Swin-V2-L~\cite{liu2021swin} and HTS-AT~\cite{chen2022hts} as enhanced alternatives in Table~\ref{tab:Enhanced_backbone} for video and audio, respectively. Additionally, in Table~\ref{tab:VAF}, we conduct new modality pairing experiments involving video, audio, and optical flow, where the backbones used are Swin-V2-L, HTS-AT, and SlowOnly~\cite{feichtenhofer2019slowfast}, respectively. As the source dataset for the optical flow modality is not provided for both pretraining and downstream tasks, we use the TV-L1~\cite{zach2007duality} algorithm for optical flow extraction to ensure data consistency. $\lambda_{1}, \lambda_{2}, \lambda_{3}, \lambda_{4}$ set to 1, 2, 1, and 1, respectively.



\subsection{Performance Analysis}
In the tables below, \textbf{bold} numbers indicate the best results, V, A, T and F represent Video, Audio, Text, and Optical Flow, respectively.

\noindent
\textbf{Open-set Cross Modal Generalization: }As shown in Table~\ref{tab:oscmg}, we compared our proposed MICU model with the previous SOTA multimodal unified representation models on the newly introduced OSCMG task. It can be observed that MICU significantly outperforms the previous SOTA models in 11 of the most important HOS metrics. The only exception is the Split2 HOS metric for VGG(a)$\rightarrow$UCF(v), where it ranks second with a value close to first place. This demonstrates the effectiveness of our proposed method on OSCMG, regardless of the dataset, its splits, or the cross-modal direction.


\noindent
\textbf{Cross Modal Generalization: } To prove that our model excels not only on the newly proposed OSCMG task, but also on the well-established CMG task, we conducted a detailed comparison with previous SOTA models. As shown in Table~\ref{tab:cmg}, MICU outperforms the previous models by a significant margin, with all 8 evaluation metrics showing clear and consistent improvements. The smallest observed improvement is as high as 2.0\%, further underscoring the robustness and superior generalizability of our approach across a wide range of tasks.


\begin{table}[ht]
    \centering
    \resizebox{0.5\textwidth}{!}{
    \begin{tabular}{ccccccccc}
    \toprule
    \centering
    Method 
                                     &
                                     \multicolumn{2}{c}{\begin{tabular}[c]{@{}c@{}}AVE\\ V$\rightarrow$A   A$\rightarrow$V\end{tabular}} & 
                                     \multicolumn{2}{c}{\begin{tabular}[c]{@{}c@{}}AVVP\\ V$\rightarrow$A   A$\rightarrow$V\end{tabular}} & 
                                     \multicolumn{2}{c}{\begin{tabular}[c]{@{}c@{}}AVE$\rightarrow$AVVP\\ V$\rightarrow$A   A$\rightarrow$V\end{tabular}} & 
                                     \multicolumn{2}{c}{\begin{tabular}[c]{@{}c@{}}UCF(v)$\leftrightarrow$VGG(a)\\ V$\rightarrow$A   A$\rightarrow$V\end{tabular}}\\
                                     \midrule
    CODIS~\citep{duan2022multi}   &36.8 & 39.7 & 32.7  & 32.6  & 40.8 & 40.6 & 50.8 & 45.2 \\ 
    TURN~\citep{zhao2022towards}    &37.6 & 39.2 & 32.4  & 32.2  & 40.6 & 41.4 & 50.4 & 46.1 \\ 
    CMCM~\citep{liu2021cross}    &46.3 & 45.8 & 36.1  & 35.2  & 47.1 & 48.2 & 51.2 & 48.3 \\ 
    DCID~\citep{xia2024achieving}    &54.1 & 55.0 & 40.4  & 40.8  & 53.0 & 52.4 & 67.1 & 60.6 \\ 
    MICU   &\textbf{56.1} & \textbf{57.1} & \textbf{45.2} & \textbf{48.2} & \textbf{56.3} & \textbf{54.9} & \textbf{75.3} & \textbf{64.5} \\ 
    \bottomrule
    \end{tabular}}
    \caption{Comparison of our model with previous SOTA models on CMG.}
    \label{tab:cmg}
\end{table}

\noindent
\textbf{Cross Modal Zero-shot Retrieval: }As shown in Table~\ref{tab:retrieval}, we also conducted Zero-shot Retrieval on two tasks, V$\leftrightarrow$T and A$\leftrightarrow$T, to demonstrate that our model still maintains an advantage in the unified representation of other modalities.


\begin{table}[ht]
    \centering
    \resizebox{0.5\textwidth}{!}{
    \begin{tabular}{cccccccc}
    \toprule
    \centering
    \multirow{2}{*}{Method} & \multicolumn{3}{c}{MSCOCO(V$\leftrightarrow$T)} & & \multicolumn{3}{c}{Clotho(A$\leftrightarrow$T)} 
    \\
     & R@1 & R@5 & R@10 & & R@1 & R@5 & R@10  \\
    \toprule
    CMCM~\citep{liu2021cross} & 0.50 & 4.20 & 7.20 & & 1.62 & 8.04 & 14.87  \\
    DCID~\citep{xia2024achieving} & 0.80 & \textbf{5.00} & 8.30 & & 2.06 & 9.00 & 16.70   \\
    MICU & \textbf{1.30} & \textbf{5.00} & \textbf{8.80} & & \textbf{2.44} & \textbf{10.96} & \textbf{18.95}   \\
    \bottomrule
    \end{tabular}}
    \caption{Comparison of our model with previous SOTA models on Zero-shot Retrieval.}
    \label{tab:retrieval}
\end{table}

\noindent
\textbf{Experiments with stronger backbones: }As shown in Table~\ref{tab:Enhanced_backbone}, all models exhibit significant performance improvements with enhanced backbones. However, under the same backbone settings, our proposed MICU consistently maintains a clear advantage, further demonstrating the effectiveness of our approach.

\begin{table}[]
    \centering
    \resizebox{0.48\textwidth}{!}{
    \begin{tabular}{ccccccccccc}
    \toprule
    \multirow{3}{*}{Dataset} & \multirow{3}{*}{Method} &\multicolumn{4}{c}{Split1} && \multicolumn{4}{c}{Split2}\\ \cline{3-6} \cline{8-11}
                             & & \multicolumn{2}{c}{Original} & \multicolumn{2}{c}{Enhanced} && 
                             \multicolumn{2}{c}{Original} & \multicolumn{2}{c}{Enhanced} \\ 
                             
                            & & V$\rightarrow$A & A$\rightarrow$V& V$\rightarrow$A & A$\rightarrow$V& &
                            
                            V$\rightarrow$A & A$\rightarrow$V& V$\rightarrow$A & A$\rightarrow$V
                            
                            
                            \\ 
                            \midrule
    \multirow{3}{*}{\bf AVE}
    &CMCM & 45.17 & 36.45 & 48.09 & 40.63& &  44.56 & 40.98 & 48.61 & 45.79 \\ 
    &DCID & 51.54 & 38.36 & 54.03 & 41.65& &  51.44 & 45.37 & 54.97 & 50.27 \\ 
    &MICU & \textbf{54.39} & \textbf{45.43} & \textbf{57.76} & \textbf{48.42} & & \textbf{59.08} & \textbf{49.17} & \textbf{61.57} & \textbf{53.13} 
 \\ \midrule
    \multirow{3}{*}{\bf UCF} 
    &CMCM & 30.00 & 34.03 & 32.72 & 35.62& &  26.79 & 29.83 & 30.25 & 32.13 \\ 
    &DCID & 34.46 & 38.09 & 35.81 & 40.15& &  28.18 & 33.65 & 30.54 & 37.18 \\ 
    &MICU & \textbf{39.82} & \textbf{39.92} & \textbf{41.27} & \textbf{42.31} & & \textbf{34.64} & \textbf{35.41} & \textbf{36.83} & \textbf{39.01} 
 \\ \midrule
    \multirow{3}{*}{\bf UCF(vf)-VGG(a)} 
    &CMCM & 72.54 & 53.45 & 76.42 & 60.20& &  62.99 & 56.87 & 66.36 & 59.98  \\ 
    &DCID & 83.58 & 65.28 & 87.69 & 68.61& &  69.87 & \textbf{62.49} & 74.23 & 65.01  \\ 
    &MICU & \textbf{87.09} & \textbf{69.69} & \textbf{90.62} & \textbf{74.64} & & \textbf{75.62} & 61.17 & \textbf{77.10} & \textbf{65.74} \\
    \bottomrule
    \end{tabular}
    }
    \caption{Comparison with previous SOTA methods on OSCMG, evaluated using HOS. Original and Enhanced refer to respective backbones in Implementation Details.}
    \label{tab:Enhanced_backbone}
\end{table}

\noindent
\textbf{Experiments with more modality combinations: }As shown in Table~\ref{tab:VAF}, all experiments are conducted using enhanced backbones, where each value represents the average result of two generalization directions. For example, $V\leftrightarrow A$ denotes the mean of $V\rightarrow A$ and $A\rightarrow V$. Our method consistently maintains a clear advantage in tasks involving optical flow, demonstrating its adaptability beyond specific modality settings. Additionally, $V\leftrightarrow F$ achieves the best overall performance, likely due to the inherent similarity between video (V) and optical flow (F) modalities.

\begin{table}[]
    \centering
    \resizebox{0.48\textwidth}{!}{
    \begin{tabular}{ccccccccc}
    \toprule
    \multirow{2}{*}{Dataset} & \multirow{2}{*}{Method}  &\multicolumn{3}{c}{Split1} & &\multicolumn{3}{c}{Split2} \\
    \cline{3-5} \cline{7-9}                        & &
                            V$\leftrightarrow$A & V$\leftrightarrow$F & A$\leftrightarrow$F & &
                            V$\leftrightarrow$A & V$\leftrightarrow$F & A$\leftrightarrow$F 
                            \\ 
                            \midrule
    \multirow{3}{*}{\bf AVE}
    &CMCM &43.73 & 45.16 & 40.98 && 47.61 & 47.97 & 34.62 \\ 
    &DCID &45.90 & 48.49 & 44.21 && 51.87 & 52.94 & 50.78 \\ 
    &MICU &\textbf{52.15} & \textbf{54.28} & \textbf{52.09} && \textbf{56.96} & \textbf{58.63} & \textbf{55.43}
 \\ \midrule
    \multirow{3}{*}{\bf UCF} 
    &CMCM &33.09 & 34.64 & 32.68 && 29.63 & 31.45 & 27.35\\ 
    &DCID &37.16 & 38.47 & 35.26 && 33.15 & 36.34 & 33.81 \\ 
    &MICU &\textbf{40.58} & \textbf{41.92} & \textbf{39.81} && \textbf{36.48} & \textbf{38.17} & \textbf{35.59}
 \\ \midrule
    \multirow{3}{*}{\bf UCF(vf)-VGG(a)} 
    &CMCM &66.37 & 68.03 & 64.70 && 60.56 & 62.89 & 60.17 \\ 
    &DCID &78.41 & 79.24 & 78.05 && 66.40 & 69.45 & 65.89 \\ 
    &MICU &\textbf{81.59} & \textbf{82.25} & \textbf{80.26} && \textbf{70.24} & \textbf{71.57} & \textbf{70.44}
 \\ \bottomrule
    \end{tabular}
    }
    \caption{Comparison with previous SOTA methods on OSCMG, evaluated using HOS. The experimental modalities include Video (V), Audio (A), and Optical Flow (F).}
    \label{tab:VAF}
\end{table}

\noindent
\textbf{Jigsaw Puzzles: }We conducted additional discussions on Jigsaw Puzzles, focusing on experiments with "without Jigsaw Puzzles," MMJP~\cite{dong2024towards} using a 6-segment split, and our proposed CUJP with 2, 4, and 8-segment splits. The limitation on the number of segments is due to the unified representation features being 256-dimensional, so the number of splits must evenly divide 256, which leads CUJP to use 2, 4, and 8 splits. In contrast, MMJP requires the features of all three modalities to be split simultaneously, which results in a multiplication factor of 3. For instance, if each modality has 2 splits, MMJP will use 6 segments. However, if each modality has 4 splits, MMJP would require $12!$ factorial permutations, which our experiments showed resulted in excessively long computation times. Therefore, MMJP is limited to 6 segments in this study.

\begin{figure}[th]
  \centering
   \includegraphics[width=1.0\linewidth]{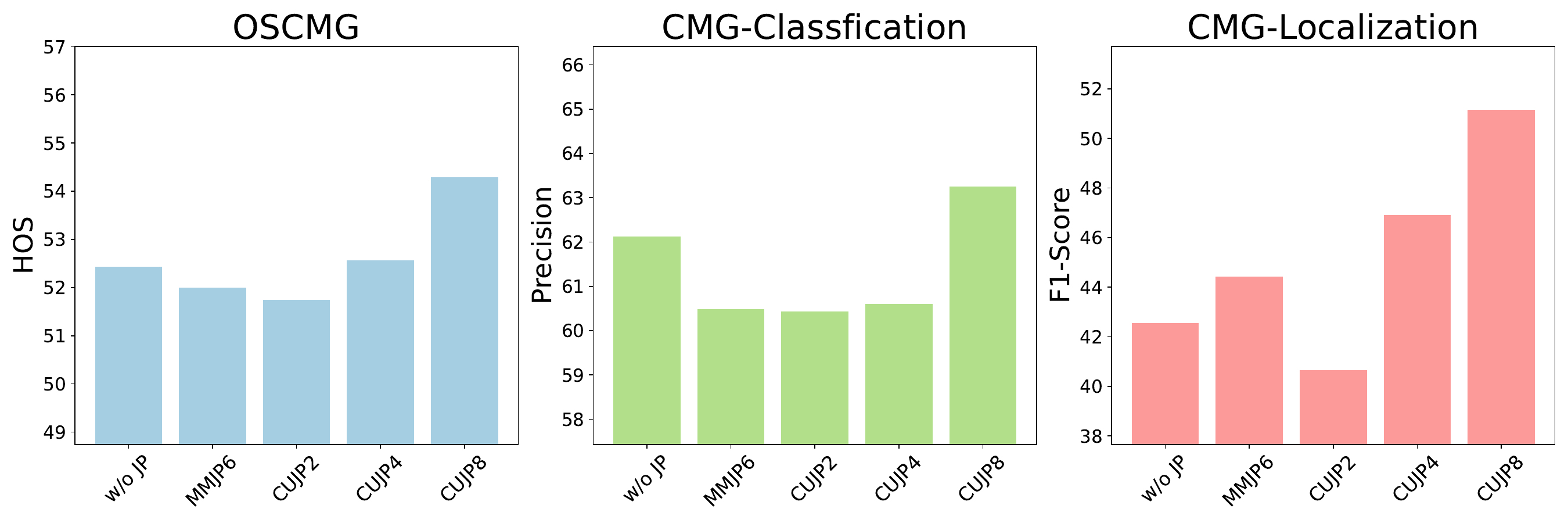}
   \caption{Experimental results of different Jigsaw Puzzles.
   \label{fig:ablation_jp}
   }
\end{figure}

\begin{table*}[htbp]
    \centering
    \resizebox{1.0\textwidth}{!}{
    \begin{tabular}{ccccccccccccccccccc}
    \toprule
    \multirow{3}{*}{Dataset} & \multirow{3}{*}{$L_{\text{fine}}$} & \multirow{3}{*}{$L_{\text{coarse}}$} & \multirow{3}{*}{$L_{\text{cujp}}$} 
                            & \multicolumn{7}{c}{Split1} & & \multicolumn{7}{c}{Split2} \\ \cline{5-11} \cline{13-19}
                            &&& & \multicolumn{3}{c}{V$\rightarrow$A} & & \multicolumn{3}{c}{A$\rightarrow$V} & & \multicolumn{3}{c}{V$\rightarrow$A} &  &\multicolumn{3}{c}{A$\rightarrow$V} \\ \cline{5-7} \cline{9-11} \cline{13-15} \cline{17-19}
                            &&& & OS* & UNK & \textbf{HOS} & & OS* &  UNK & \textbf{HOS} & & OS* & UNK & \textbf{HOS} & &OS* & UNK & \textbf{HOS} \\
                            \midrule
    \multirow{7}{*}{\bf AVE}  
    &\checkmark &-&- & 8.07&10.61&9.17& &7.17&13.97&9.48 & & 4.43&26.74&7.60& &5.06&16.28&7.72 \\ 
    &-&\checkmark&- & 45.29&54.75&49.57& &29.15&65.92&40.42 & & 38.61&83.72&52.85& &26.90&81.40&40.43 \\ 
    &-&-&\checkmark & 7.17&28.49&11.46& &7.62&35.75&12.57 & & 5.38&19.77&8.46& &5.38&36.05&9.36 \\ 
    &\checkmark&\checkmark&- & 49.33 &59.78 &54.05 & &40.39 &51.19 &45.15 & & 43.04 &77.91 &55.45 & &36.73 &72.09 &48.67 \\
    &\checkmark&-&\checkmark & 7.17 &12.23 &41.34 & &0.90 &99.44 &1.78 & & 5.38 &17.44 &8.22 & &5.70 &1.16 &1.93 \\
    &-&\checkmark&\checkmark & 43.50 &71.51 &54.09 & &35.87 &47.58 &40.90 & & 45.57 &63.47 &53.05 & &30.38 &48.84 &37.46 \\
    &\checkmark&\checkmark&\checkmark & 51.57 & 57.54 & \textbf{54.39} & & 34.98 & 64.80 & \textbf{45.43} & & 47.15 & 79.07 & \textbf{59.08} & & 35.44 & 80.23 & \textbf{49.17} \\ \midrule
    \multirow{7}{*}{\bf UCF} 
    &\checkmark &-&- & 4.28&23.54&7.24& &7.08&4.12&5.21 & & 2.16&20.74&3.92& &3.53&0.82&1.32 \\ 
    &-&\checkmark&- & 24.86&62.11&35.51& &30.76&48.86&37.75 & & 20.25&56.75&29.85& &23.86&56.57&33.56 \\ 
    &-&-&\checkmark & 5.85&20.96&9.15& &6.12&24.59&9.80 & & 2.90&18.43&5.00& &3.53&18.98&5.95 \\ 
    &\checkmark&\checkmark&- & 29.06 &60.30 &39.22 & &27.10 &68.69 &38.87 & & 24.25 &56.07 &33.86 & &27.93 &52.63 &\textbf{36.49} \\
    &\checkmark&-&\checkmark & 8.39 &20.08 &11.83 & &6.95 &8.98 &7.83 & & 3.96 &20.79 &6.65 & &2.66 &18.93 &4.67 \\
    &-&\checkmark&\checkmark & 28.53 &58.89 &38.44 & &30.23 &58.56 &39.88 & & 22.45 &66.08 &33.52 & &24.96 &63.90 &35.90 \\
    &\checkmark&\checkmark&\checkmark & 29.40 & 61.69 & \textbf{39.82} & &    27.48 & 72.96 & \textbf{39.92} & & 24.33 & 60.05 & \textbf{34.64}   & &  23.90 & 68.25 & 35.41 \\ \midrule
    \multirow{7}{*}{\bf UCF(v)$\leftrightarrow$VGG(a)} 
    &\checkmark &-&- & 13.47&30.05&18.60& &0.17&96.80&0.34 & & 9.04&25.71&13.38& &1.61&98.43&3.16 \\ 
    &-&\checkmark&- & 70.80&91.79&79.94& &60.95&70.23&65.26 & & 60.40&84.94&70.60& &45.24&72.93&55.84 \\ 
    &-&-&\checkmark & 12.26&67.60&20.76& &12.07&43.65&18.91 & & 6.08&67.80&11.16& &9.00&53.92&15.42 \\ 
    &\checkmark&\checkmark&- & 79.16 &82.05 &80.58 & &65.26 &70.70 &67.87 & & 58.39 &68.34 &62.97 & &51.03 &63.76 &56.69 \\
    &\checkmark&-&\checkmark & 2.76 &86.50 &5.35 & &9.57 &51.96 &16.16 & & 13.19 &9.03 &10.72 & &5.98 &61.75 &10.90 \\
    &-&\checkmark&\checkmark & 75.86 &86.83 &80.98 & &65.26 &73.12 &68.97 & & 63.59 &81.16 &71.31 & &54.88 &69.13 &61.19 \\
    &\checkmark&\checkmark&\checkmark & 81.72 & 93.23 & \textbf{87.09} & & 68.71 & 70.70 & \textbf{69.69} & & 66.77 & 87.18 & \textbf{75.62} & & 47.43 & 86.13 & \textbf{61.17} \\ \bottomrule
    \end{tabular}
    }
    \caption{Ablation study of the three losses proposed by our model on OSCMG.}
    \label{tab:ablation_module_oscmg}
\vspace{-3mm}
\end{table*}

The specific experimental results are shown in Figure~\ref{fig:ablation_jp}, where we separate the classification and localization tasks of CMG into two charts to display the model differences more clearly. It can be observed that MMJP6 performs worse than w/o jp in both OSCMG and CMG-Classification, showing improvements only in CMG-Localization. In contrast, CUJP's performance improves as the number of splits increases, showing a clear upward trend, with CUJP8 significantly outperforming all other configurations. Additionally, CUJP4 already consistently outperforms MMJP6, which demonstrates that for tasks related to multimodal unified representations, the CUJP setup is more suitable.

\noindent
\textbf{Ablation Study: }Since $L_{recon}$ and $L_{commit}$ are standard losses for discrete representations and not the novelty of this paper, their effectiveness has been established in prior work. Therefore, our ablation study focuses on the newly proposed loss.

As shown in Table~\ref{tab:ablation_module_oscmg}, we conducted a detailed ablation study on the three newly proposed losses in the MICU architecture, namely $L_{fine}$, $L_{coarse}$, and $L_{cujp}$. First, by observing the first three rows for each dataset, it is evident that $L_{coarse}$ is the foundation of the model, as without it, a unified representation cannot be constructed. This is apparent because $L_{coarse}$ represents contrastive learning of overall semantics, and without overall semantics, a representation space cannot be built. Next, comparing the 2nd and 4th rows, it can be observed that the combination of $L_{coarse}$ and $L_{fine}$ further improves the model's performance, with noticeable gains in 11 HOS metrics. This indicates that $L_{fine}$ provides fine-grained temporal knowledge that $L_{coarse}$ alone cannot learn, helping the model construct a more refined representation space. Similarly, the comparison between the 2nd and 6th rows also shows improvements in 11 HOS metrics, indicating that $L_{cujp}$ also helps build a better representation space, with the modality-agnostic Jigsaw Puzzles proving to be highly effective. The 5th row shows the same effect as the 2nd row, confirming that without contrastive learning of overall semantics, a representation space cannot be constructed. The 7th row demonstrates that the combination of all three components achieves the optimal result.

\noindent
\textbf{Additional Experiments: }Further experiments, including the mask setting of FCMI (Sec~\ref{sec:mask}), codebook size hyperparameter selection (Sec~\ref{sec:codebooksize}), ablation study on CMG (Sec~\ref{sec:ablation_cmg}), computational efficiency analysis (Sec~\ref{sec:com_eff}), and visualization of the discrete representation space (Sec~\ref{sec:visual}), are provided in the supplementary material.

\section{Conclusion}
To advance the evaluation of multimodal unified representations in complex scenarios, we introduce the Open-set Cross-Modal Generalization (OSCMG) task, which specifically addresses the challenges of open-set detection and multimodal alignment. To tackle these challenges, we propose the MICU method, which integrates two key components: Fine-Coarse Masked Multimodal InfoNCE and Cross-Modal Unified Jigsaw Puzzle. These components offer complementary strategies, combining fine-grained masked contrastive learning with modality-agnostic self-supervised learning to enhance generalization and alignment across diverse modalities. Our approach achieves state-of-the-art performance on the OSCMG task and demonstrates significant improvements over previous models on the CMG task. Overall, we introduce a novel task to evaluate the performance of multimodal unified representations in open-set domains, and propose a new method to effectively address the challenges posed by this task.


\section*{Acknowledgments}
This work was supported by National Key R\&D Program of China (2022ZD0162000) and National Natural Science Foundation of China (62222211).
{   
    \small
    \bibliographystyle{ieeenat_fullname}
    \bibliography{main}
}
\clearpage
\setcounter{page}{1}
\maketitlesupplementary

The citation numbers are consistent with those in the main text.

\section{Mask of FCMI}
\label{sec:mask}
\begin{figure}[th]
  \centering
   \includegraphics[width=1.0\linewidth]{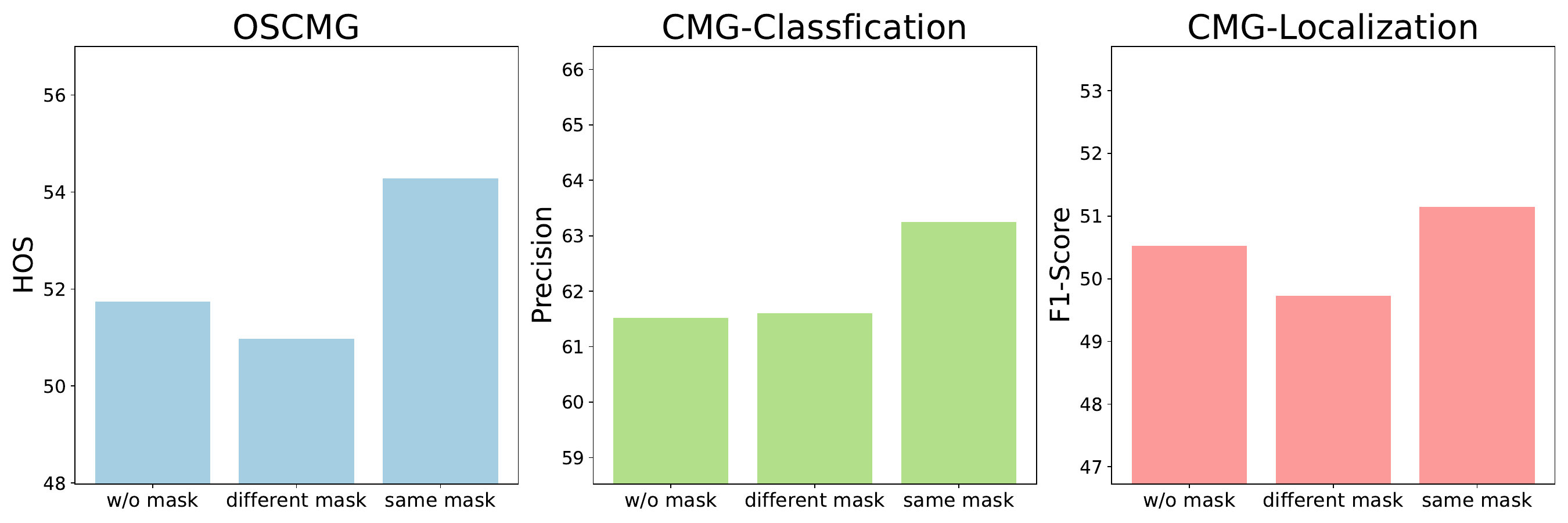}
   \caption{Experimental results of different Mask.
   \label{fig:ablation_mask}
   }
\end{figure}

We also conducted an analysis on different masking strategies. As shown in Figure~\ref{fig:ablation_mask}, applying the same mask to paired multimodal samples helps improve model performance. This approach facilitates more precise and detailed alignment between modalities, ensuring semantic consistency in the unmasked regions while applying the mask to the same positions across modalities. In contrast, using different masking positions for each modality in paired samples leads to a decline in performance, as it disrupts the semantic alignment across the modalities.

\vspace{-3mm}
\section{Codebook Size}
\label{sec:codebooksize}
\begin{figure}[th]
  \centering
   \includegraphics[width=1.0\linewidth]{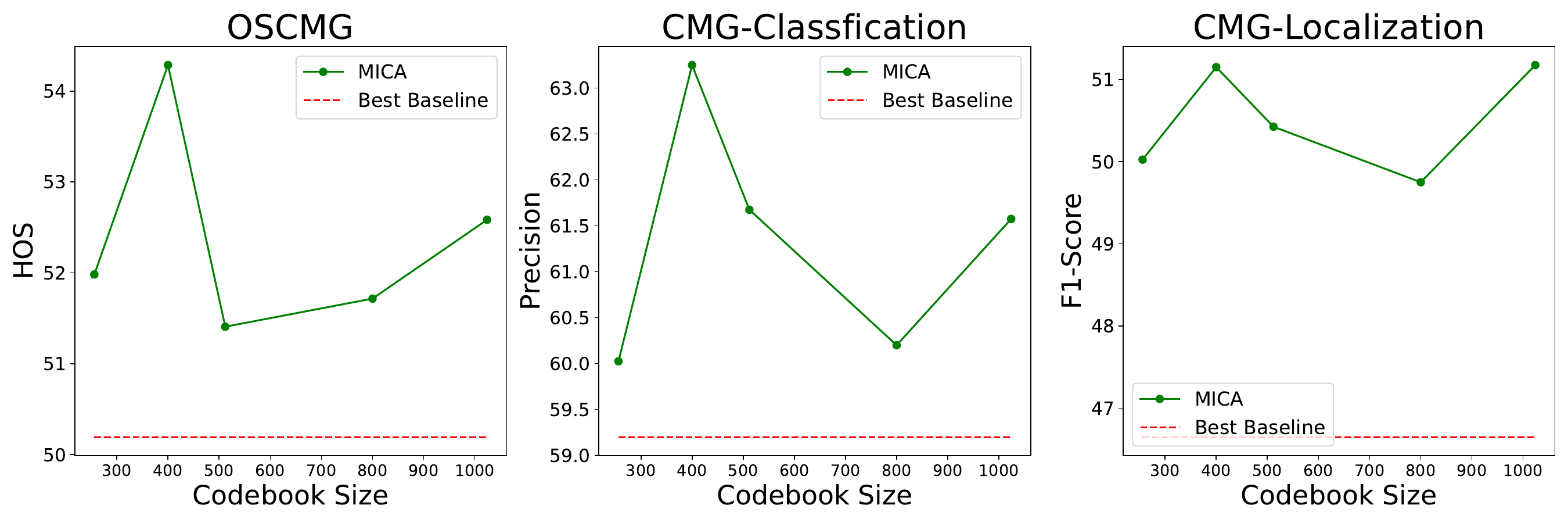}
   \caption{Experimental results of different Codebook Size.
   \label{fig:ablation_codebooksize}
   }
\end{figure}

The size of the representation space also affects the model's performance. As shown in Figure~\ref{fig:ablation_codebooksize}, we experimented with five different settings: 256, 400, 512, 800, and 1024. Among these, 400 led by a significant margin over the other settings. Therefore, we chose a codebook size of 400 as the final setting for our model.

\vspace{-3mm}
\section{Ablation on CMG}
\label{sec:ablation_cmg}
The experimental results of Table~\ref{tab:ablation_module_cmg} and Table~\ref{tab:ablation_module_oscmg} are similar. $L_{coarse}$ serves as the foundation of the model, while $L_{fine}$ and $L_{cujp}$ further refine the unified representation space and enhance the model's open-domain detection capabilities.

\begin{table}[ht]
    \centering
    \resizebox{0.48\textwidth}{!}{
    \begin{tabular}{ccccccccccc}
    \toprule
    \centering
    \multirow{1}{*}{$L_{\text{fine}}$} & \multirow{1}{*}{$L_{\text{coarse}}$} & \multirow{1}{*}{$L_{\text{cujp}}$}                                                                      
                                     &
                                     \multicolumn{2}{c}{\begin{tabular}[c]{@{}c@{}}AVE\\ V$\rightarrow$A   A$\rightarrow$V\end{tabular}} & 
                                     \multicolumn{2}{c}{\begin{tabular}[c]{@{}c@{}}AVVP\\ V$\rightarrow$A   A$\rightarrow$V\end{tabular}} & 
                                     \multicolumn{2}{c}{\begin{tabular}[c]{@{}c@{}}AVE$\rightarrow$AVVP\\ V$\rightarrow$A   A$\rightarrow$V\end{tabular}} & 
                                     \multicolumn{2}{c}{\begin{tabular}[c]{@{}c@{}}UCF(v)$\leftrightarrow$VGG(a)\\ V$\rightarrow$A   A$\rightarrow$V\end{tabular}}\\
                                     \midrule
    \checkmark&-&-  &7.1 & 5.2 & 13.4  & 13.7  & 15.9 & 7.4 & 10.5 & 8.2 \\ 
    -&\checkmark&-  &54.3 & 55.2 & 39.6  & 37.8  & 50.5 & 46.3 & 70.3 & 61.7 \\ 
    -&-&\checkmark  &5.6 & 5.1 & 0  & 6.0  & 0 & 0 & 13.0 & 9.7 \\ 
    \checkmark&\checkmark&-   &\textbf{56.1} & 57.0 & 38.9  & 35.8  & 52.2 &43.3 & 70.8 & \textbf{64.6} \\ 
    \checkmark&-&\checkmark   &6.4 &4.8 & 13.4  & 13.7 &15.9& 7.4 & 11.1 & 8.2 \\ 
    -&\checkmark&\checkmark    &53.8 & 52.4 & 43.8  & 45.9  & \textbf{56.7} & \textbf{54.9} & 67.4 & 62.3 \\ 
    \checkmark&\checkmark&\checkmark  &\textbf{56.1} & \textbf{57.1} & \textbf{45.2} & \textbf{48.2} & 56.3 & \textbf{54.9} & \textbf{75.3} & 64.5 \\ 
    \bottomrule
    \end{tabular}
    }
    \caption{Ablation study of the three losses proposed by our model on CMG.}
    \label{tab:ablation_module_cmg}
\vspace{-3mm}
\end{table}

\section{Computational Efficiency}
\label{sec:com_eff}
As shown in Table~\ref{t2}, compared to CMCM~\cite{liu2021cross} and DCID~\cite{xia2024achieving}, our method requires more GPU memory and longer per-epoch training time, but achieves better performance, reflecting a trade-off between performance and resources. CUJP8, despite having more split block reordering, optimizes memory usage and reduces training time compared to MMJP6~\cite{dong2024towards}. Increasing the number of splits (CUJP4 vs. CUJP8) leads to higher memory usage but better performance in multimodal alignment. CMCM requires more epochs due to warm-start techniques. Inference time differences across all models are minimal and task-dependent. For reproducibility, the complete source code is provided in the supplementary materials.

\begin{table}[h]
\centering
\resizebox{0.48\textwidth}{!}{
\begin{tabular}{l|c|c|c|c|c}
\hline
\textbf{Method} & \textbf{GPU Memory Usage} & \textbf{Time per Epoch} & \textbf{Total Epochs}& \textbf{OSCMG Avg.} & \textbf{CMG Avg.} \\
\hline
CMCM & 6.25GB & 1.41h & 8 & 44.47 & 44.78 \\
DCID & 7.90GB & 1.72h & 5 & 50.19 & 52.93 \\
MICU (MMJP6) & 14.77GB & 2.30h & 5 & 52.00 & 52.46 \\
MICU (CUJP4) & 9.07GB & 2.13h & 5 & 52.56 & 53.75 \\
MICU (CUJP8) & 13.30GB & 2.22h & 5 & 54.29 & 57.20 \\
\hline
\end{tabular}
}
\caption{Comparison of computational efficiency with the original backbone (batch size: 80, GPU: RTX 3090).}
\label{t2}
\end{table}

\vspace{-5mm}
\section{Unified Representation Space Visualization}
\label{sec:visual}
\begin{figure*}
  \centering
  \begin{subfigure}{0.48\linewidth}
    \includegraphics[width=1.0\linewidth]{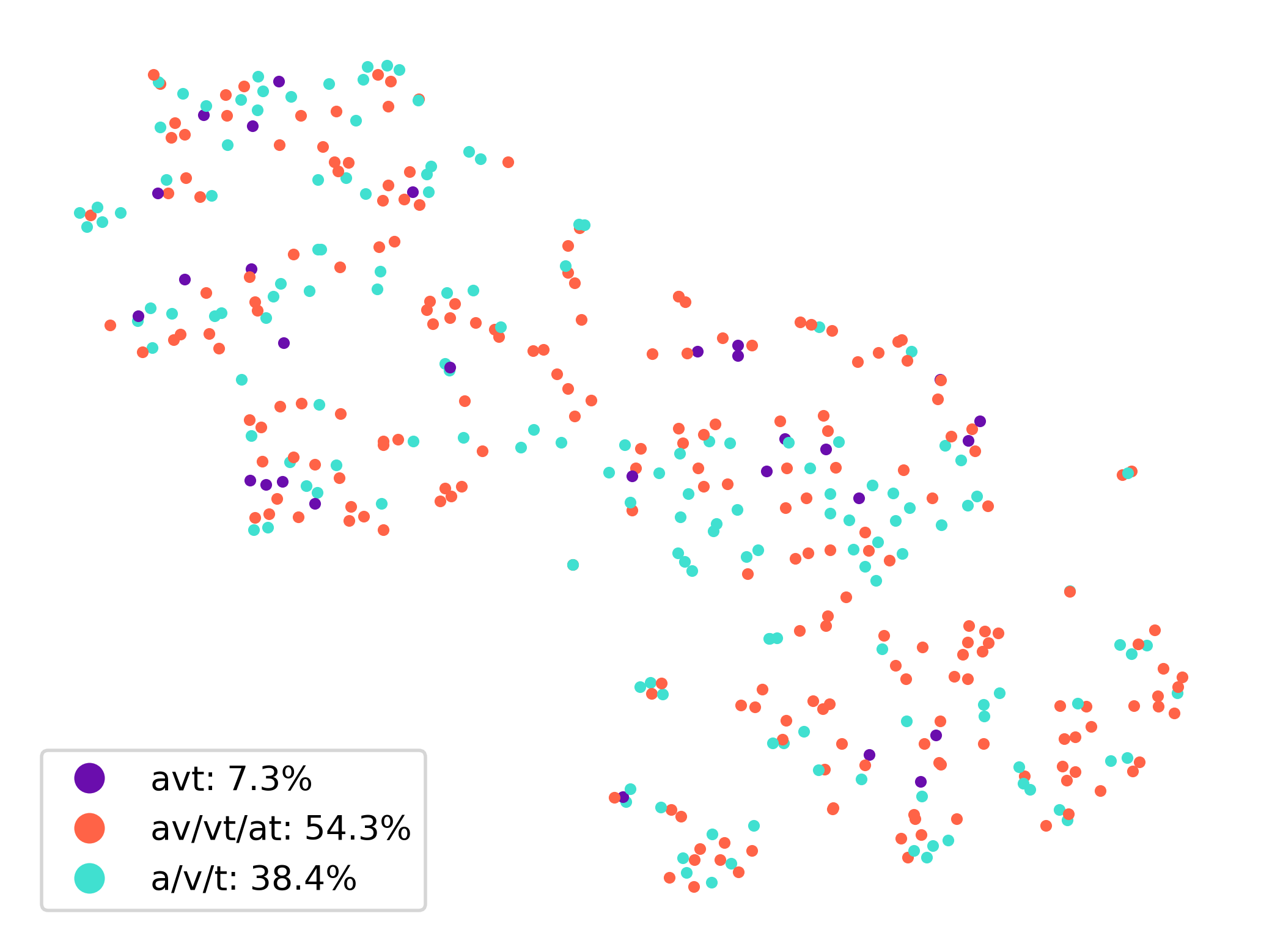}
    \caption{DCID Representation Space Visualization}
    \label{fig:short-a}
  \end{subfigure}
  \hfill
  \begin{subfigure}{0.49\linewidth}
    \includegraphics[width=1.0\linewidth]{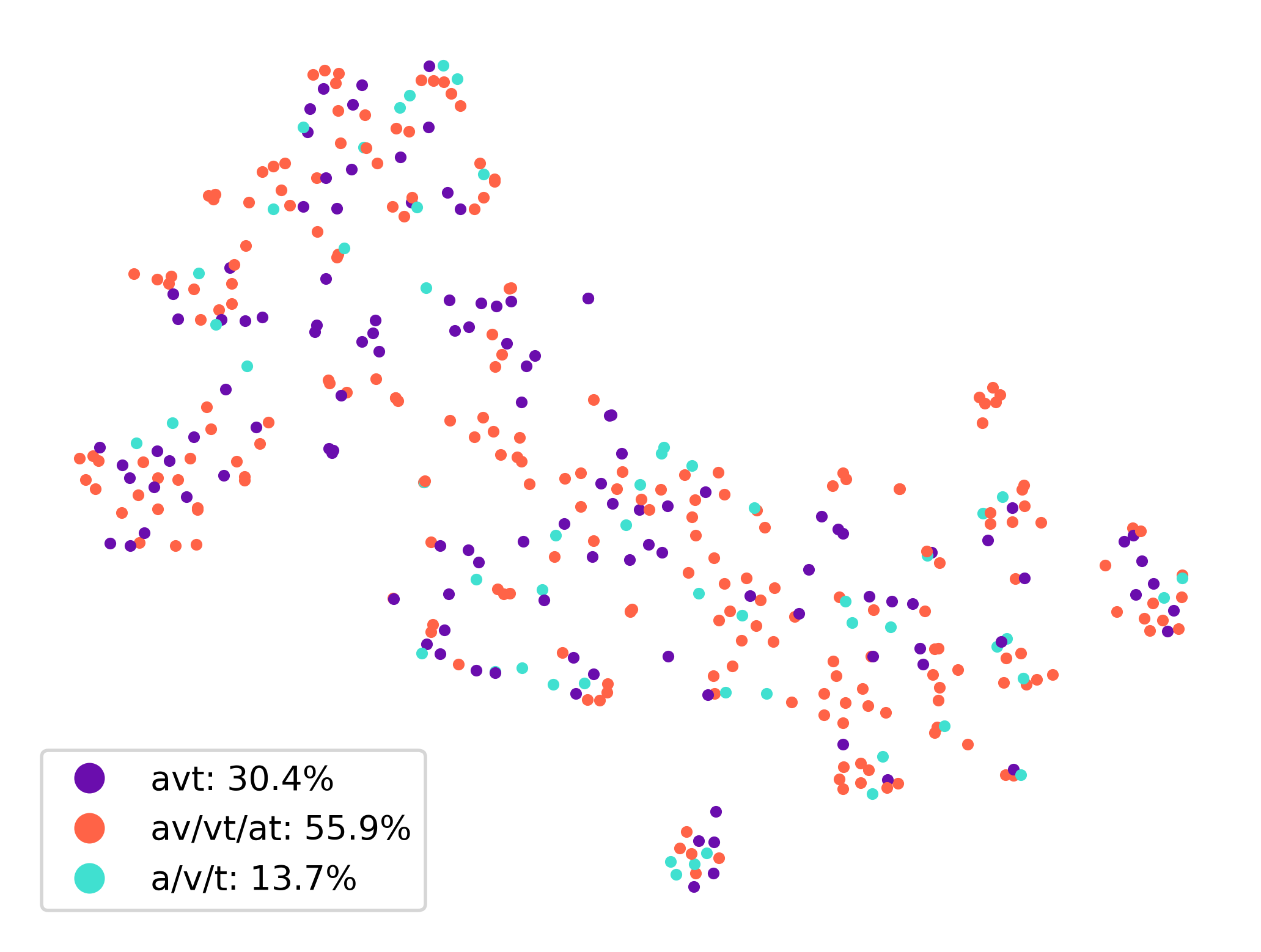}
    \caption{MICU Representation Space Visualization}
    \label{fig:short-b}
  \end{subfigure}
  \caption{Purple (avt) indicates where all three modalities have quantized activations $\geq$10\%, orange (av/vt/at) for two modalities, and cyan (a/v/t) for a single modality.}
  \label{fig:codebook_visualization}
\end{figure*}

As shown in Figure~\ref{fig:codebook_visualization}, the two subfigures illustrate the representation spaces of DCID~\cite{xia2024achieving} after pre-training and our proposed model. The visualization maps audio-video-text triplets from the Valor32K dataset~\cite{chen2023valor} into the unified representation space (codebook). Codewords quantized by all three modalities with a proportion of $\geq$10\% are marked in purple, those shared by any two modalities with $\geq$10\% appear in orange, while those dominated by a single modality are shown in cyan. The bottom left of the figure indicates the proportion of each color.

A higher proportion of cyan suggests an imbalanced multimodal distribution, indicating larger modality discrepancies, whereas more purple signifies stronger cross-modal alignment, aligning with the goal of a unified representation. As observed, our model achieves significantly better multimodal integration compared to DCID.


\nocite{fang2024ace, wang2025irbridge, wang2025towards, ji2024wavtokenizer,sinkd,sink2,cui2025streetsurfgs,cui2025layoutenc,optical}


\end{document}